\documentclass[conference, a4paper]{IEEEtran}
\IEEEoverridecommandlockouts
\usepackage{cite}
\usepackage{amsmath,amssymb,amsfonts}
\usepackage{algorithmic}
\usepackage{graphicx}
\usepackage{textcomp}
\usepackage{xcolor}
\usepackage{caption}
\usepackage{subcaption}
\usepackage{booktabs}

\DeclareMathOperator{\argmax}{arg\,max}

\def\BibTeX{{\rm B\kern-.05em{\sc i\kern-.025em b}\kern-.08em
    T\kern-.1667em\lower.7ex\hbox{E}\kern-.125emX}}
\begin{document}

\title{Decentralized Multi-Robot Formation Control \\ Using Reinforcement Learning
\thanks{This work is supported by European Union's Horizon Europe research program Widening participation and spreading excellence, through project Strengthening Research and Innovation Excellence in Autonomous Aerial Systems (AeroSTREAM) under grant agreement no. 101071270, and by European Union's Horizon 2020 research and innovation program Future and Emerging Technologies, through project Smart Biohybrid Phyto-Organisms for Environmental In Situ Monitoring (WATCHPLANT) under grant agreement no. 101017899. The work of doctoral student Marko Križmančić has been supported in part by the “Young researchers’ career development project—training of doctoral students” of the Croatian Science Foundation funded by the European Union from the European Social Fund.}
\thanks{Research work presented in this article has been supported by the project "Razvoj autonomnog besposadnog višenamjenskog broda" project (KK.01.2.1.02.0342) co-financed by the European Union from the European Regional Development Fund within the Operational Program "Competitiveness and Cohesion 2014-2020". The content of the publication is the sole responsibility of the project partner UNIZG-FER.}
}


\author{\IEEEauthorblockN{1\textsuperscript{st} Juraj Obradovi\'c}
\IEEEauthorblockA{\textit{University of Zagreb, Faculty of} \\ \textit{Electrical Engineering and Computing}\\
Zagreb, Croatia \\
juraj.obradovic@fer.hr}
\and
\IEEEauthorblockN{2\textsuperscript{nd} Marko Kri\v{z}man\v{c}i\'c}
\IEEEauthorblockA{\textit{University of Zagreb, Faculty of} \\ \textit{Electrical Engineering and Computing}\\
Zagreb, Croatia \\
marko.krizmancic@fer.hr}
\and
\IEEEauthorblockN{3\textsuperscript{rd} Stjepan Bogdan}
\IEEEauthorblockA{\textit{University of Zagreb, Faculty of} \\ \textit{ Electrical Engineering and Computing}\\
Zagreb, Croatia \\
stjepan.bogdan@fer.hr}
}

\maketitle

\begin{abstract}
This paper presents a decentralized leader-follower multi-robot formation control based on a reinforcement learning (RL) algorithm applied to a swarm of small educational Sphero robots. Since the basic Q-learning method is known to require large memory resources for Q-tables, this work implements the Double Deep Q-Network (DDQN) algorithm, which has achieved excellent results in many robotic problems.
To enhance the system behavior, we trained two different DDQN models, one for reaching the formation and the other for maintaining it. The models use a discrete set of robot motions (actions) to adapt the continuous nonlinear system to the discrete nature of RL.
The presented approach has been tested in simulation and real experiments which show that the multi-robot system can achieve and maintain a stable formation without the need for complex mathematical models and nonlinear control laws.
\end{abstract}


\section{Introduction}
\label{sec:intro}
Formation control is a part of many multi-robot applications that have drawn significant attention in the last decade. The robot formations are used to perform a collective task, such as the transport of objects~\cite{AlonsoMora2017}, monitoring~\cite{Saska2017}, search and rescue~\cite{Arnold2018}, and exploration~\cite{Baranzadeh2017}, while in the same time forming and keeping a specific shape (i.e. relative distances among members of the group). Through the years, many different approaches have been implemented to achieve that goal~\cite{Scharf2004}. 

In \textit{leader-follower}~\cite{Wang1999} approach one agent represents a leader with a static control law, usually following some predefined trajectory or navigating to a set goal. The rest of the group are followers that track the position of the leader with a specific offset needed to form a desired shape. The leader can either be virtual~\cite{Su2009} or one of the robots in the group. \textit{Behaviour-based}~\cite{Balch1998} approaches take inspiration from swarms in nature, such as flocks of birds and schools of fish. Each agent in the swarm exhibits several simple desired behaviors: cohesion, alignment, separation from its neighbors, obstacle avoidance, etc.~\cite{Reynolds1987}. The agent's momentary movement is then calculated as the mean of all desired behaviors at that moment. In \textit{virtual structure}~\cite{Low2011, miklic} approach, the whole formation is seen as one structure. The structure and position of the robots within it are defined. The robot's speed is then calculated to track the movement of the formation.

Classical methods for implementing these common approaches often require complex mathematical models~\cite{Lin2014} and ultimately yield non-linear control laws which are difficult to implement. That is why in this paper we present a decentralized leader-follower multi-robot formation control developed using a reinforcement learning (RL) algorithm and apply it to a swarm of small educational Sphero robots. Reinforcement learning algorithms embody an appealing idea of learning by interacting with the environment, a very natural way of learning that has significant applicability in multi-robot systems.
Instead of carefully designing the control law needed for stable and smooth movement of the formation and handling different scenarios such as obstacle avoidance and intrusion detection, we can expose our agents to the expected environment and teach them the optimal behavior. 

Many different RL algorithms have shown good results in the domain of formation control. 
Knopp \textit{et al.} used the GQ($\lambda$) algorithm in combination with the leader-follower approach and applied it to a group of e-puck robots~\cite{Knopp2017}. Based on the measured angle and distance errors from the robot in front, the trained model successfully maintained a worm-like formation in the simulation. However, the real-world experiments didn't perform that well due to the limitations of the robots.
In~\cite{Sadhukhan2021} Proximal Policy Optimization (PPO) algorithm was adopted for the formation control in a simulated environment with obstacles. 
The authors compare two reward models, one based on the actions of individual agents, and one based on the actions of the team as a whole. Using the centralized policy trained with those models, the formation of robots successfully reached the goal in 5\% and 7\% of the cases, respectively.
Much better results were obtained in~\cite{Zhou2019}. Zhou \textit{et al.} presented a complete solution for controlling a group of mobile robots in cluttered environments. They utilized a convolutional neural network for end-to-end localization and designed a novel actor-critic algorithm called Momentum Policy Gradient (MPG). In all of the experiments, the tracking error was less than a few centimeters. The only downside of the method is that it is centralized, and therefore less likely to be scalable.

The authors in~\cite{He2020} take a different path and demonstrate the effectiveness of a hybrid approach based on Deep Deterministic Policy Gradient (DDPG). Instead of directly controlling the formation, the RL algorithm was used to learn the optimal gain parameters for the classical non-linear formation control method. The presented results were excellent, but the method is applicable only to the specific case with two robots.
Recently, Xie \textit{et al.} proposed an improved version of the DDPG algorithm by combining it with the advantage function which helps to select better actions and expedite the learning process (ADDPG) ~\cite{Xie2021}. The algorithm is applied to a chain leader-follower approach in the control of unmanned surface vehicles. In oppose to previous algorithms, state space does not need to be discretized, and the procedure is completely decentralized as trained policies run independently on each vehicle, bringing the method closer to real-world practicality.

This work implements the Double Deep Q-Networks (DDQN) algorithm which showed excellent results in different robotic problems like steering controller design~\cite{Wu2019} and obstacle avoidance~\cite{Xue2019}. It has also been used in formation control~\cite{Sui2018}, but in a centralized manner and only applicable to teams consisting of two robots. In our application, we have opted for the leader-follower approach due to its robustness, simplicity, and the fact that it is the perfect approach for implementing decentralized control. The leader agent is given a predefined trajectory to follow using a classical PID-based controller, and followers locally use the trained DDQN model to select actions that will track the leader's motion. Thus, followers only require bearing and distance measurements to the leader and potential obstacles in contrast to knowing the complete formation.

The contributions of this paper are manifold:
\begin{enumerate}
    \item We present the DDQN model for decentralized formation reaching, formation maintenance, and obstacle avoidance in multi-robot systems based on the leader-follower approach.
    \item We show that this method can be used to achieve stable formation movements using only a simple mathematical model of the robot and without the need for a sophisticated non-linear controller.
    \item By presenting successful simulation and real experiments with Sphero robots, we show that the model trained entirely in simulation can easily and effectively bridge the "sim-to-real" gap. 
\end{enumerate}

The remainder of the paper is structured as follows. First, in Section~\ref{sec:background} we give a short theoretical background. In Section~\ref{sec:system} we provide the problem description and give a detailed overview of our proposed system. Simulation results and real experiments are described in Section~\ref{sec:results}. Finally, a short conclusion and description of future work are given in Section~\ref{sec:conclusion}.

\section{Theoretical background}
\label{sec:background}

\begin{figure}[tb]
    \centering
    \includegraphics[width=1.0\columnwidth]{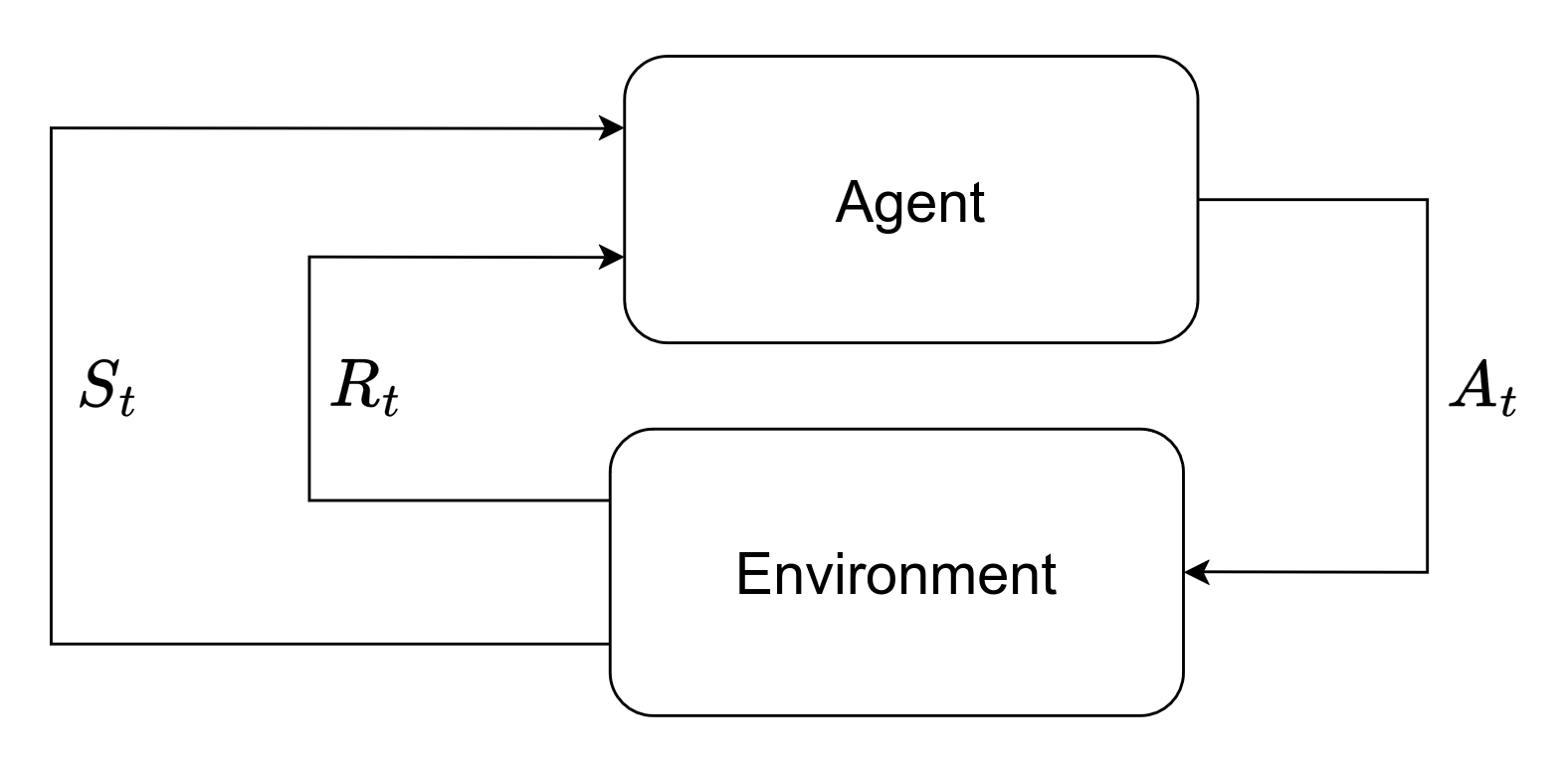}
    \caption{Markov decision process graphical model \cite{Sutton2018}}
    \label{fig:mdp}
\end{figure}

Reinforcement learning (RL) is a machine learning technique in which an agent (in this case a robot) is trying to find the optimal behavior by interacting with the environment and maximizing some objective specified as a reward. This type of problem is most commonly described as Markov Decision Process (MDP). In MDP, the system is divided into two subsystems, agent and environment. At each time-step $t$, the agent that is in some state $s_t \in S$, chooses to perform action $a_t \in A$, based on current behavior policy $\pi(a|s)$. Depending on the problem, $S$ and $A$ can be either discrete or continuous. After taking the selected action, the environment "responds" by moving the agent to a new state $s_{t+1}$ and producing the reward $r_{t+1}$. This process, shown in Fig. \ref{fig:mdp}, repeats until the agent reaches a terminal state or makes the maximum allowed number of steps.

One of the simplest, and yet one of the most popular RL algorithms is the model-free, off-policy Q-learning algorithm \cite{Watkins1992}. The policy in Q-learning is coded as a \textit{state-action Q-table} and it is iteratively updated using Bellman's dynamic programming equation:

\begin{multline}
    \mathbf{Q}_{k+1}(\mathbf{s}_{k},a_{k}) \leftarrow \mathbf{Q}_k(\mathbf{s}_{k},a_{k}) + \\ + \alpha \cdot \left[ r_{k} + \gamma \cdot \max_{a}\mathbf{Q}_k(\mathbf{s}_{k+1}, a) - \mathbf{Q}_k(\mathbf{s}_{k},a_{k}) \right]
     \label{eq:bellman_dp_q} 
\end{multline}

Unfortunately, due to the limited memory and generalization properties of Q-tables, the basic Q-learning algorithm is inadequate for larger problems such as formation control. To overcome its flaws, Mnih \textit{et al.} decided to approximate state-action values using deep neural networks, resulting in the Deep Q-Networks (DQN) algorithm \cite{Mnih2015}.

However, like the Q-learning, this algorithm is known to overestimate action values in stochastic environments due to the maximization step in the iterative update \cite{Hasselt2016}. The problem occurs because it is using the same samples to estimate action values $q(a, s)$ and determine the best action to take in the next step \cite{Sutton2018}. One way to mitigate this is to use two estimates ($Q_A$, $Q_B$), forming the Double Deep Q-Networks (DDQN) algorithm \cite{Hasselt2016}. One estimator $Q_A$ can then be used to calculate the maximum (i.e. best) action $a^* = \argmax_a(Q_A)$ and $Q_B$ can be used to provide estimates of that value $Q_B(a^*) = Q_B(\argmax_a(Q_A))$. Likewise, this process can be reversed to calculate $Q_A(a^*) = Q_A(\argmax_a(Q_B))$. If the two estimators work with different sets of data, the algorithm produces the policy closer to optimum in fewer episodes \cite{Sutton2018}.
\section{The system overview}
\label{sec:system}
Implementation of the leader-follower approach in this paper consists of two main parts. The first part is a simple PID controller that moves the lead agent in a specific pattern with a constant speed. We assume this speed is less than the maximum speed of the followers. The second part is the trained reinforcement learning model that moves the agent to the predefined position relative to the leader and maintains it during motion. Formations can then be created by applying this reinforcement learning model to multiple agents with different relative positions to the leader or chaining them to produce more complex structures.

\subsection{Sphero robots}
\begin{figure}[tb]
    \centering
    \includegraphics[width=0.45\columnwidth]{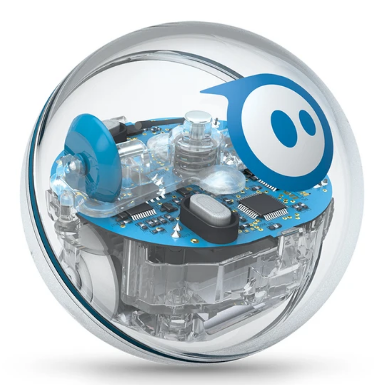}
    \caption{Sphero SPRK+ \cite{sphero}}
    \label{fig:sphero}
\end{figure}

To demonstrate formation control in a real-world scenario we use Sphero SPRK+ educational spherical robots \cite{sphero}. As presented in Fig.~\ref{fig:sphero}, the Sphero robot consists of a transparent plastic sphere, two wheels driven by DC motors, two passive wheels for stability, and an electronic board with modules for sensing, communication, and control. Users can send the commanded velocity and direction from the PC or mobile phone using Bluetooth 4.0 protocol. The onboard microprocessor uses an accelerometer, gyroscope, and motor encoders to stabilize the internal structure and track the commanded velocity specified in the x and y directions of the global reference frame. 

In line with keeping the controller much simpler than the classical non-linear controllers, we model our agents with single-integrator dynamics:
\begin{equation}
    \dot{\mathbf{x}}_i = \mathbf{a}_i
\end{equation}
where $\mathbf{x}_i = [x_i, y_i]^T$ is the Cartesian position of the agent $i$, and $\mathbf{a}_i = [u_{x,i},u_{y,i}]^T$ are actions of the controller.

\subsection{State and action space}
The action space of the controller is discrete. The length of the velocity vector $\| \mathbf{a}_i \|$ is kept constant and slightly larger than the movement speed of the lead agent, whereas the direction is chosen from eight equally-distanced angles around the robot,
\begin{equation}
\angle(\mathbf{a}_i) = \{ j\pi/4: j=0,1,...,7 \}.    
\end{equation}

The state $\mathbf{s}_i$ is defined as a tuple $\langle \mathbf{l}_i, \mathbf{t}_i,  \mathbf{o_1}_i, \mathbf{o_2}_i \rangle$. Each element of the tuple represents a vector between the follower $i$ and the leader, target, and two nearest obstacles respectively.

\begin{figure}[tb]
    \centering
    \includegraphics[width=0.75\columnwidth]{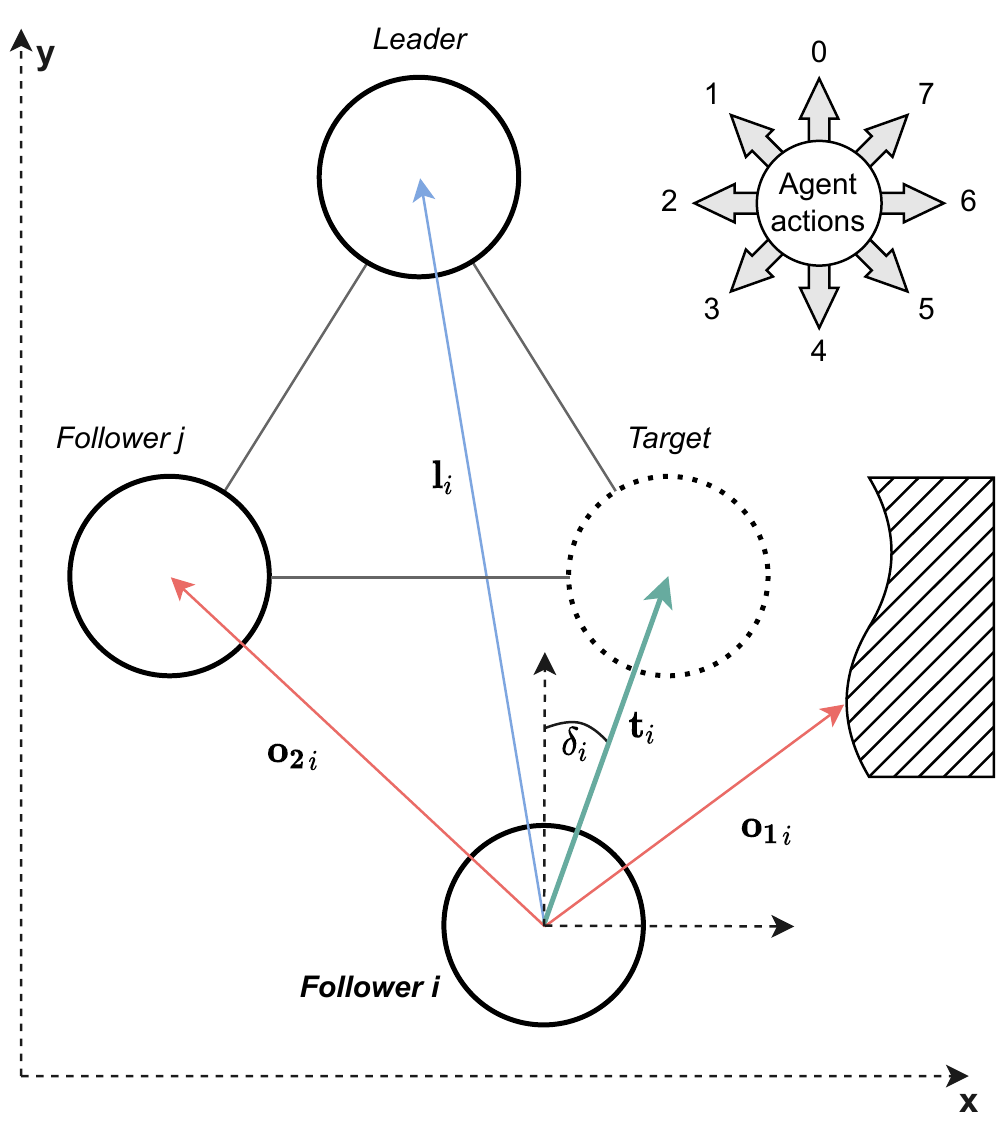}
    \caption{Graphical description of agent's states and actions}
    \label{fig:state-action}
\end{figure}

To maximally generalize the model and avoid collisions in the formation, both the leader and other followers are regarded as dynamic obstacles. For the same reason, we use the target point defined relative to the position of the leader as the reference for the follower. Definitions and relations of state and action variables are visualized in Fig.~\ref{fig:state-action}.

\subsection{Reward function}
\begin{figure}[tb]
    \centering
    \includegraphics[width=0.5\textwidth]{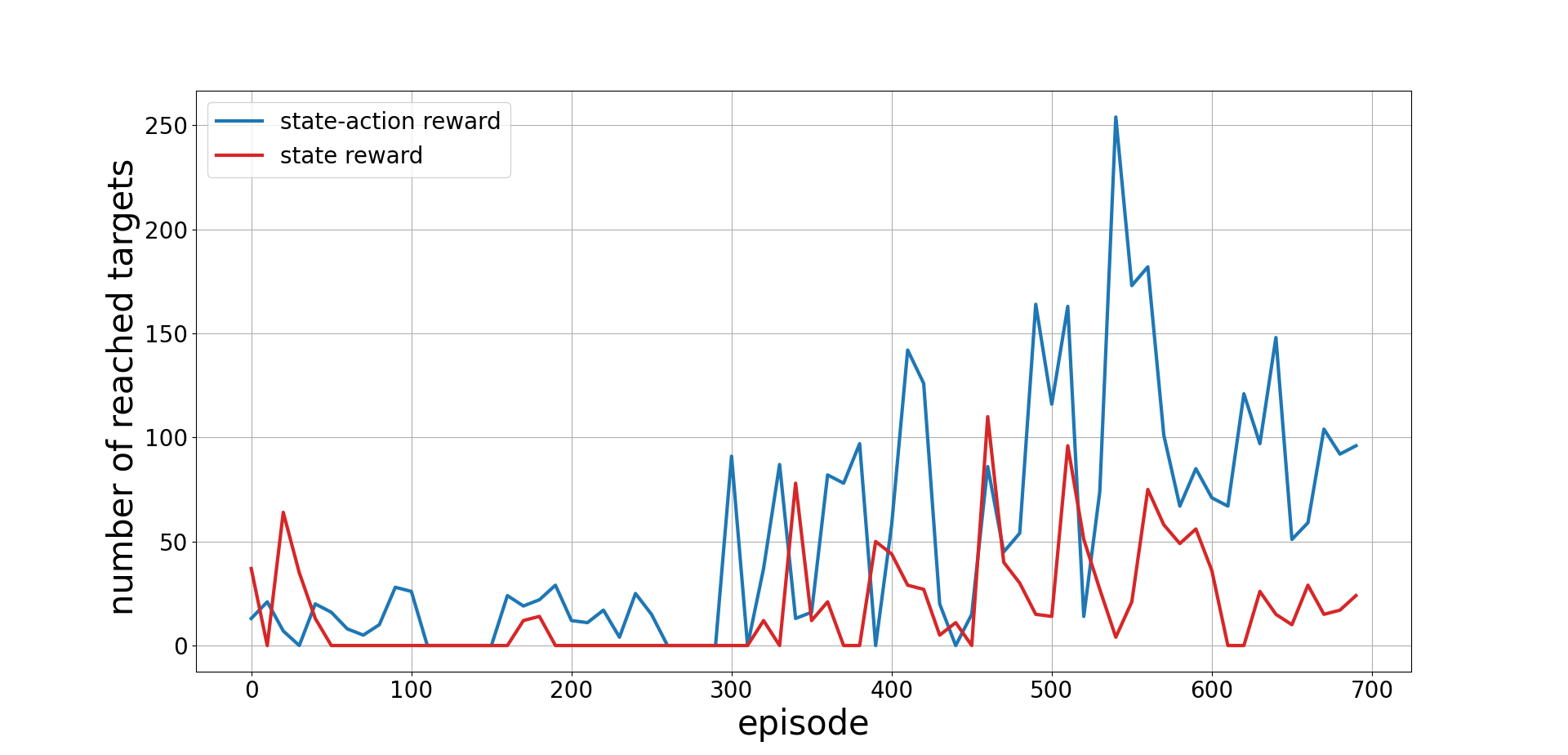}
    \caption{Comparison of state and state-action reward function. Lines represent the number of iterations a robot was within some target radius for every 10 episodes.}
    \label{fig:actionState}
\end{figure}

We tested several different reward functions, with varying success in formation reaching and formation keeping. Discrete reward functions, such as rewarding the agent when it arrives in the defined target radius, did not provide good results. Continuous reward functions, either based solely on the agent's state or considering the taken action, provide a much better environment for learning. For example, in the first approach, the agent would receive a reward proportional to its distance from the goal, while in the second, the reward is given only if it's moving in the correct direction. As we show in Fig.~\ref{fig:actionState}, including the action information in the reward function greatly improves the performance of the algorithm.

Unsurprisingly, we noticed that for simpler tasks, agents learn quickly with excellent results. Therefore, our system consists of two different models, one for formation reaching and another for formation keeping. 

The reward function for the formation-keeping model is defined as
\begin{equation}
    R^{keep}_i = r^t_i - \max(r^o_i). \label{eq:R_keep}
\end{equation}
The first term of the reward function $r^t_i$ is the reward for moving toward the target based on the agent's current distance and selected action:
\begin{equation}
   \label{eq:target}
   r^t_i = r^{td}_i \cdot r^{t\theta}_i.
\end{equation}
The distance component $r^{td}_i$ is calculated using a simple exponential function of the agent's distance from the target:
\begin{equation}
    r^{td}_i = 2^{\|\mathbf{t}_i\|}. \label{eq:distance}
\end{equation}
For the second part, we first calculate the absolute difference between the currently selected action and the angle of the target relative to the agent:
\begin{equation}
    \delta_i = \left|\angle(\mathbf{t}_i) - a^j_i\frac{\pi}{4}\right|,
\end{equation}
where $a^j_i$ is the index of the robot action as displayed in Fig.~\ref{fig:state-action}. The angle difference $\delta_i$ is normalized to range $[0, \pi]$.
Finally, $r^{t\theta}_i$ is calculated as:
\begin{equation}
    r^{t\theta}_i = 1 - \frac{\delta_i}{\pi} - \frac{n_{a}^-}{n_a}, \label{eq:angle}
\end{equation}
where $n_a=8$ is the total number of possible actions, and $n_{a}^-=5$ is the number of actions that would result in negative behavior. In other words, if the normalized absolute difference between the target direction and selected action is less than $3/8$, the reward will be positive, and otherwise, it will be negative.

The second term of the reward function is the maximization of rewards for avoiding obstacles. Similarly to the first term, it is defined as
\begin{equation}
    r^o_i =r^{od}_i \cdot r^{o\theta}_i, \label{eq:obstacle}
\end{equation}
where $r^{o\theta}_i$ is calculated the same way as (\ref{eq:angle}) but with $\angle(\mathbf{o}_i)$ instead of $\angle(\mathbf{t}_i)$, and
\begin{equation}
    r^{od}_i =2^{1/(\|\mathbf{o}_i\| + 0.1)},
\end{equation}
where $\|\mathbf{o}_i\|$ is the distance between the agent $i$ and the obstacle.
The value of $r^o_i$ is calculated for every obstacle in the environment, and only the maximum value is used in the final reward function for formation keeping defined in (\ref{eq:R_keep}).

The target-keeping model is defined in a similar way. The only difference is that once the target is reached agent should no longer deal with obstacle avoidance. Therefore, for target keeping the reward is defined as: 
\begin{equation}
    r = r_t,
\end{equation}
and $r_t$ is defined the same as in the model for target reaching (Eq. \ref{eq:target}).
    
At first, the agent uses the formation reaching model, and when the target is reached, it switches to the formation keeping. This method showed much better results than having one reward function for two different tasks.

\subsection{Neural network architecture}
In reinforcement learning, artificial networks are widely used to estimate the optimal policy. The neural network used in this paper consists of an 8-dimensional input layer, two 64-dimensional densely connected hidden layers with ReLU activation, and 8-dimensional output with linear activation. The primary task of our neural network is to suggest which action the agent should take based on its current state (given by the angle and distance values relative to the leader, target, and two obstacles). At the output layer, the neural network produces estimated q-values for each possible action from the current state. The DDQN algorithm incorporates an epsilon-greedy approach with exponentially decaying epsilon values \cite{Sutton2018}, as described in Table~\ref{tab:params}. Initially, the exploration is prioritized by setting the epsilon value to $\epsilon=1$. As training progresses, the epsilon value is updated by multiplying it with the decay factor, shifting the focus towards exploiting the learned knowledge rather than exploration.

\begin{table}[hb]
\centering
\caption{DDQN hyperparameters used in training}
\label{tab:params}
\begin{tabular}{@{}ll@{}}
\toprule
\textbf{Hyperparameter}                                                                          & \textbf{Value} \\ \midrule
Batch size                                                                                       & 64             \\
Replay memory size                                                                               & 200 000        \\
\begin{tabular}[c]{@{}l@{}}Min. number of elements in  the memory before training\end{tabular} & 100 000        \\
Max. number of steps per episode                                                                 & 300            \\
Starting value of $\epsilon$                                                                     & 1              \\
Value of $\epsilon$ decay                                                                                 & 0.9975         \\
Min. value of $\epsilon$                                                                         & 0.05           \\
Discount factor $\gamma$                                                                       & 0.99           \\
Learning rate $\alpha$                                                                         & 0.0003         \\ \bottomrule
\end{tabular}
\end{table}

Most of the final training parameters are set to values commonly used in the literature. We chose a batch size of 64 to allow for faster learning while remaining compatible with lower-end hardware. The size of the replay memory was also adjusted to suit the available computational resources. Additionally, we set the maximum number of steps per episode to 300. This value was determined empirically and represents the maximum anticipated number of steps required by the agent to reach its target. If the agent exceeds this limit, it likely deviated from the optimal path and will not reach its target within the given episode.

\subsection{Simulation environment}
For training and testing purposes, we use a 3D physically realistic open-source Gazebo simulator \cite{Gazebo} and its interface to Robot Operating System (ROS) \cite{Ros} which acts as a framework for communication between all parts of the presented system. The training setup consists of 4 Sphero robots spawned in an empty Gazebo world. One robot represents the leader tracking a predefined trajectory using a simple controller. Another robot is using the DDQN algorithm to learn appropriate actions for following the leader, while the other two simulate the obstacles. During the learning process, obstacles remain stationary, but the trained policy can also avoid dynamic obstacles.

\section{Results}
\label{sec:results}

\subsection{Simulation experiments}
\begin{figure}[t]
    \centering
    \begin{subfigure}[b]{0.48\columnwidth}
        \centering
        \includegraphics[width=\textwidth]{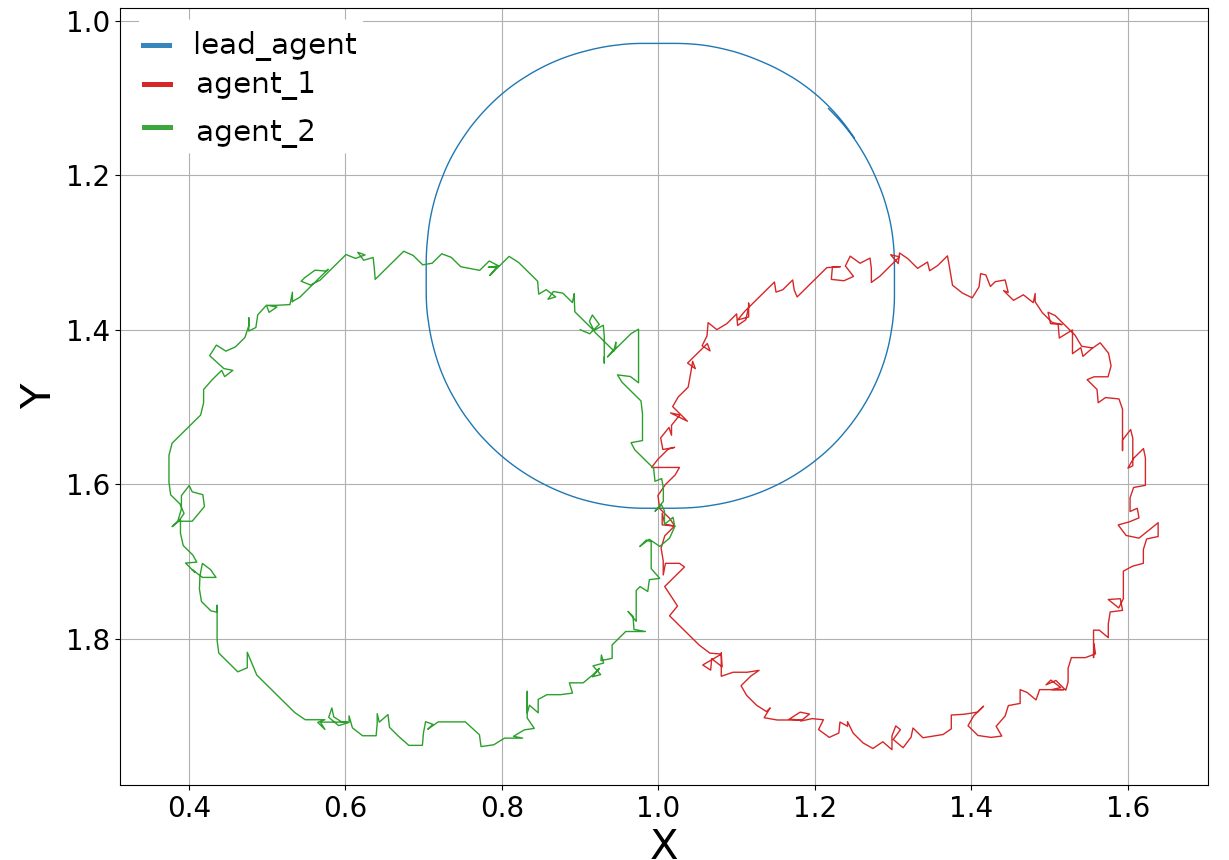}
        \caption{Leader moving in circle \\ pattern}
        \label{fig:simulation-circle}
    \end{subfigure}
    \hfill
    \begin{subfigure}[b]{0.48\columnwidth}
        \centering
        \includegraphics[width=\textwidth]{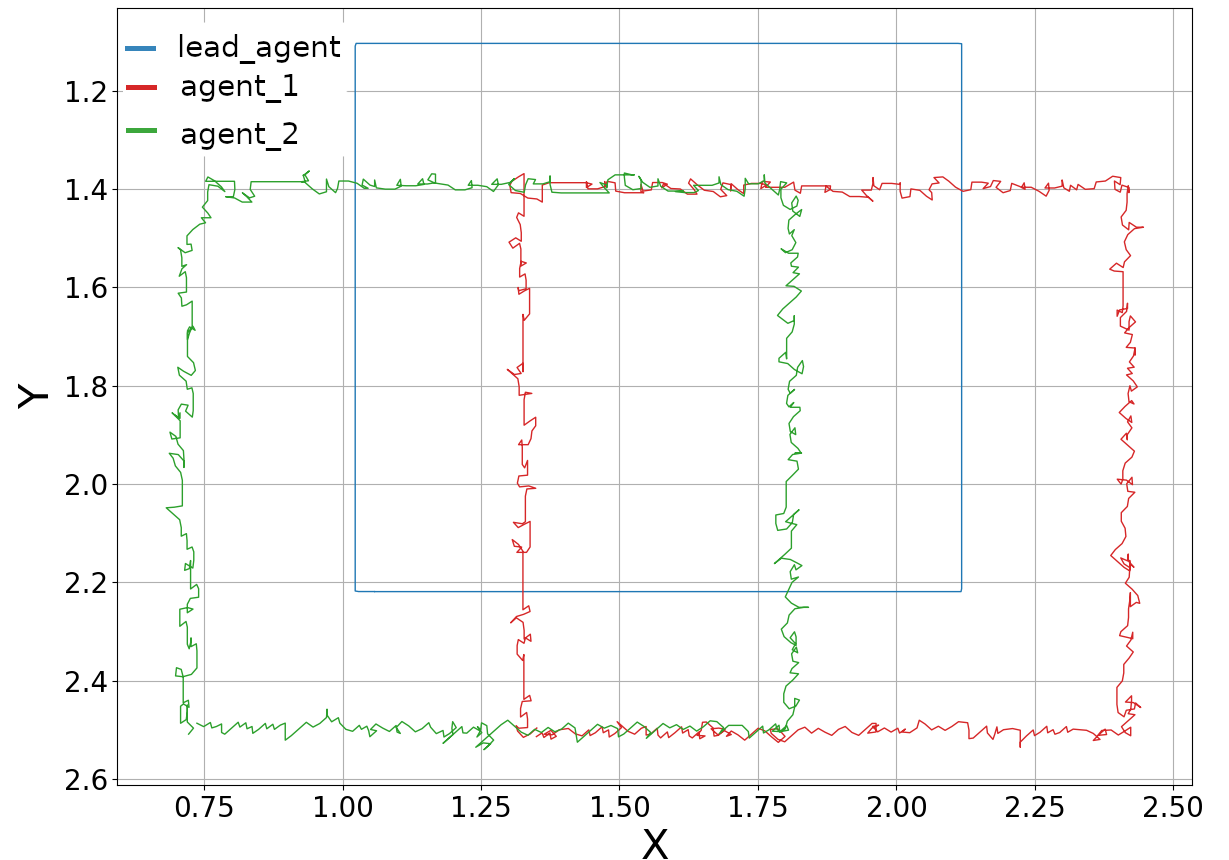}
        \caption{Leader moving in square \\ pattern}
        \label{fig:simulation-square}
    \end{subfigure}
    \caption{Simulation results for circular and square motion}
    \label{fig:simulation-results} 
\end{figure}

In the first experiment, follower agents are placed in target points of the specified formation and tasked with keeping their relative position using the formation-keeping model. The leader agent continuously moves in a specific pattern. As shown in Fig.~\ref{fig:simulation-results}, both agents successfully follow the leader's circular (Fig.~\ref{fig:simulation-circle}) and square (Fig.~\ref{fig:simulation-square}) motion. Since moving towards the target is rewarded, they tend to oscillate slightly to keep their position.

\begin{figure}[t]
    \centering
    \begin{subfigure}[b]{0.24\textwidth}
        \centering
        \includegraphics[width=\textwidth]{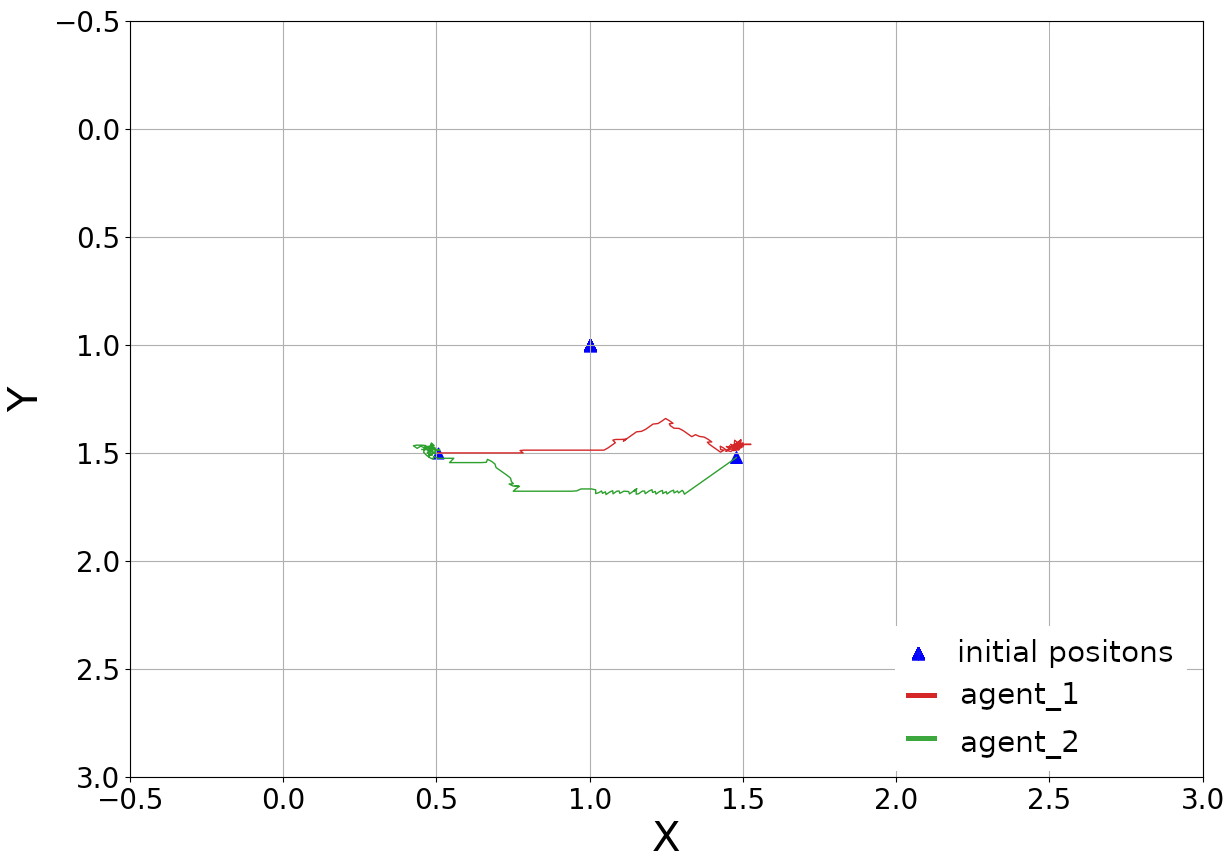}
        \caption{Initial setup 1}
        \label{fig:simulation-is1}
    \end{subfigure}
    \hfill
    \begin{subfigure}[b]{0.24\textwidth}
        \centering
        \includegraphics[width=\textwidth]{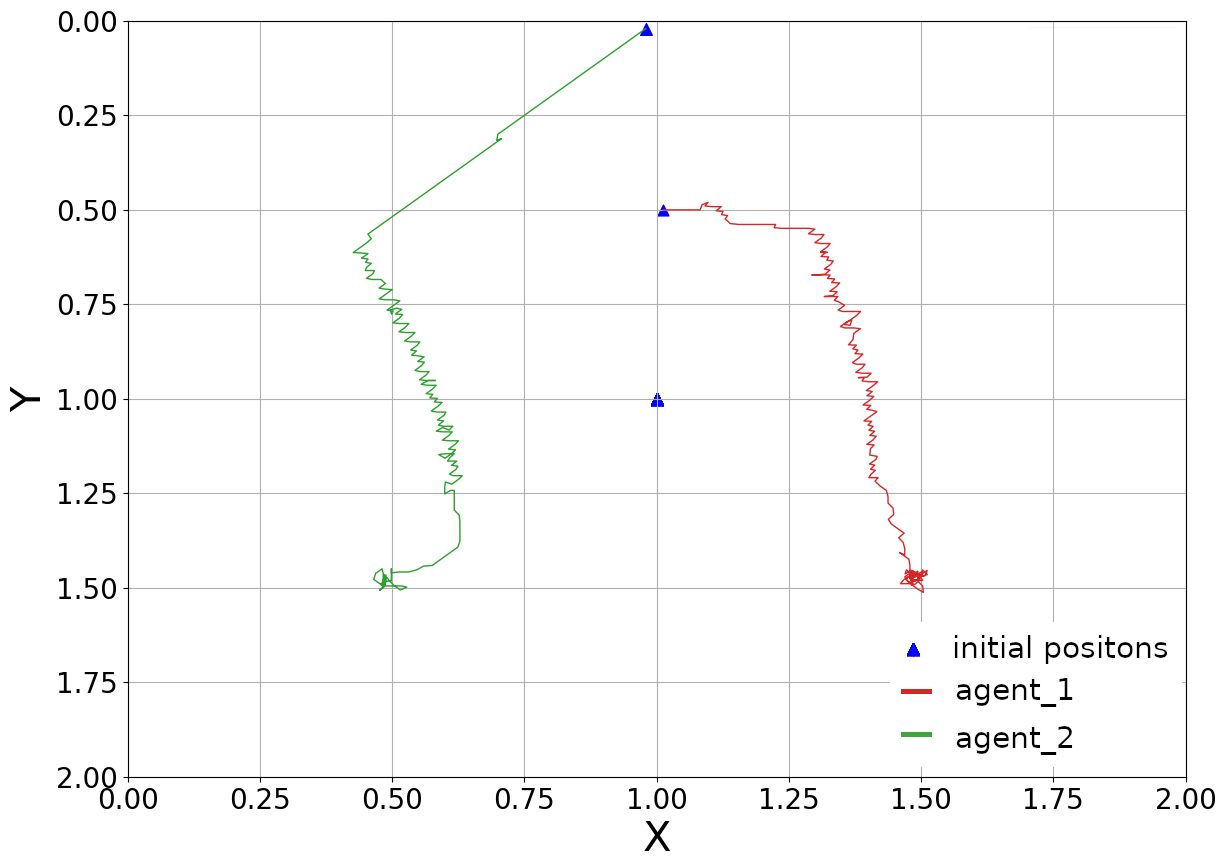}
        \caption{Initial setup 2}
        \label{fig:simulation-is2}
    \end{subfigure}
    \hfill
    \begin{subfigure}[b]{0.24\textwidth}
        \centering
        \includegraphics[width=\textwidth]{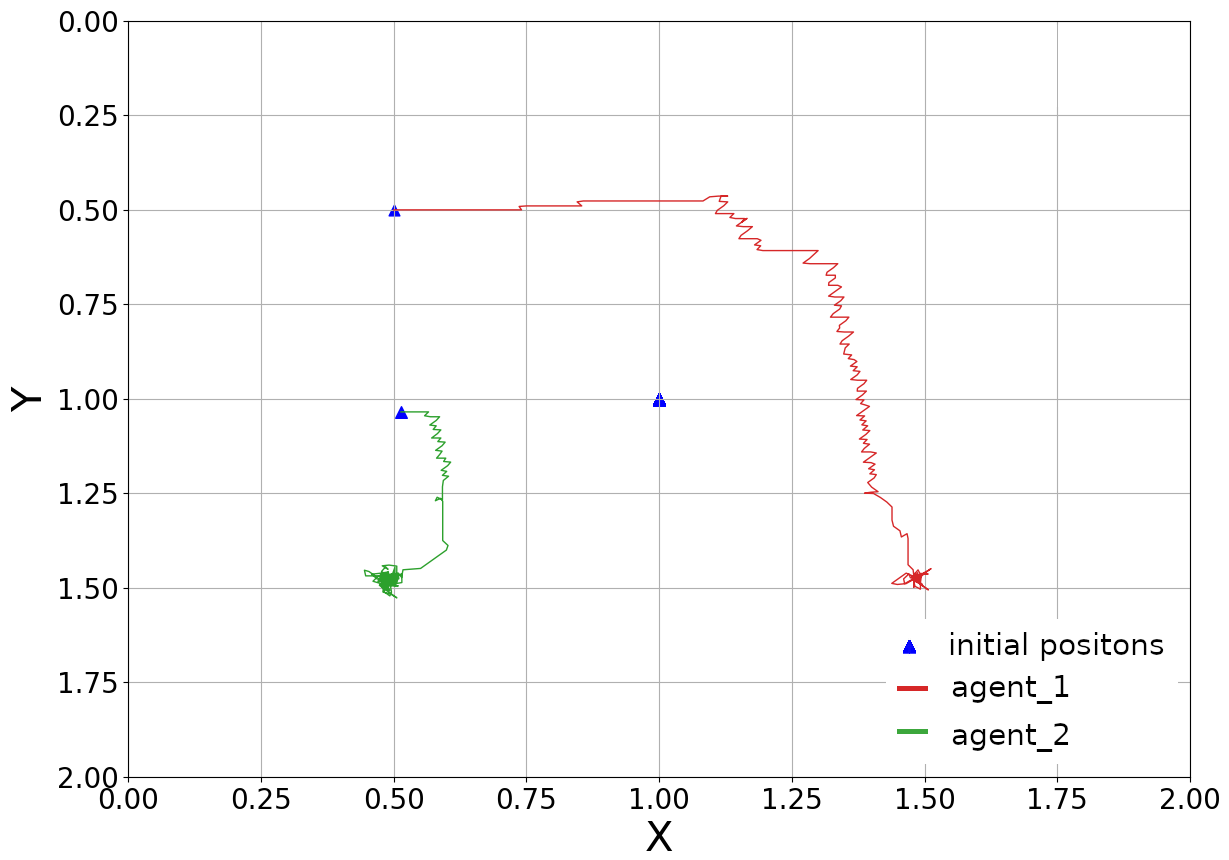}
        \caption{Initial setup 3}
        \label{fig:simulation-is3}
    \end{subfigure}
    \hfill
    \begin{subfigure}[b]{0.24\textwidth}
        \centering
        \includegraphics[width=\textwidth]{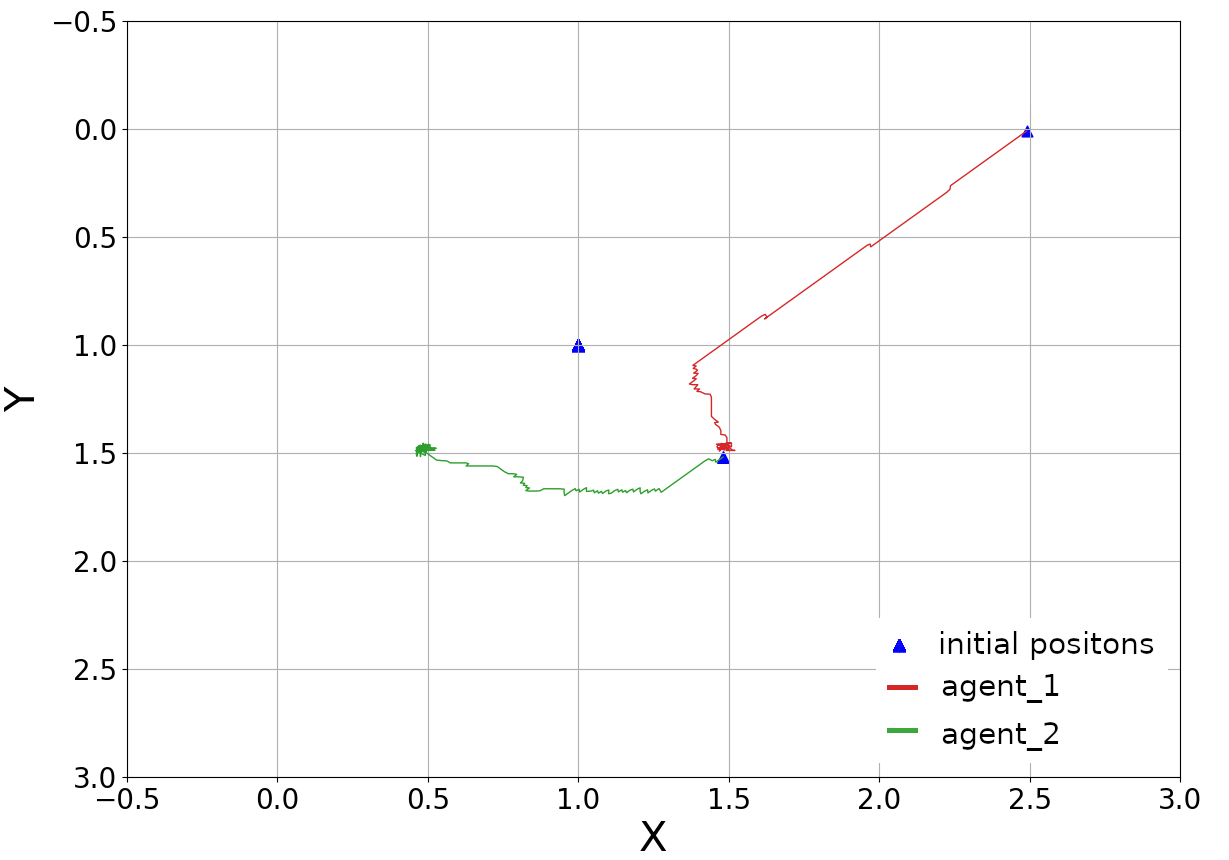}
        \caption{Initial setup 4}
        \label{fig:simulation-is4}
    \end{subfigure}
    \caption{Target reaching and keeping model with static leader}
    \label{fig:simulation-resultsReaching} 
\end{figure}

\begin{figure}[t]
    \centering
    \includegraphics[width=0.85\columnwidth]{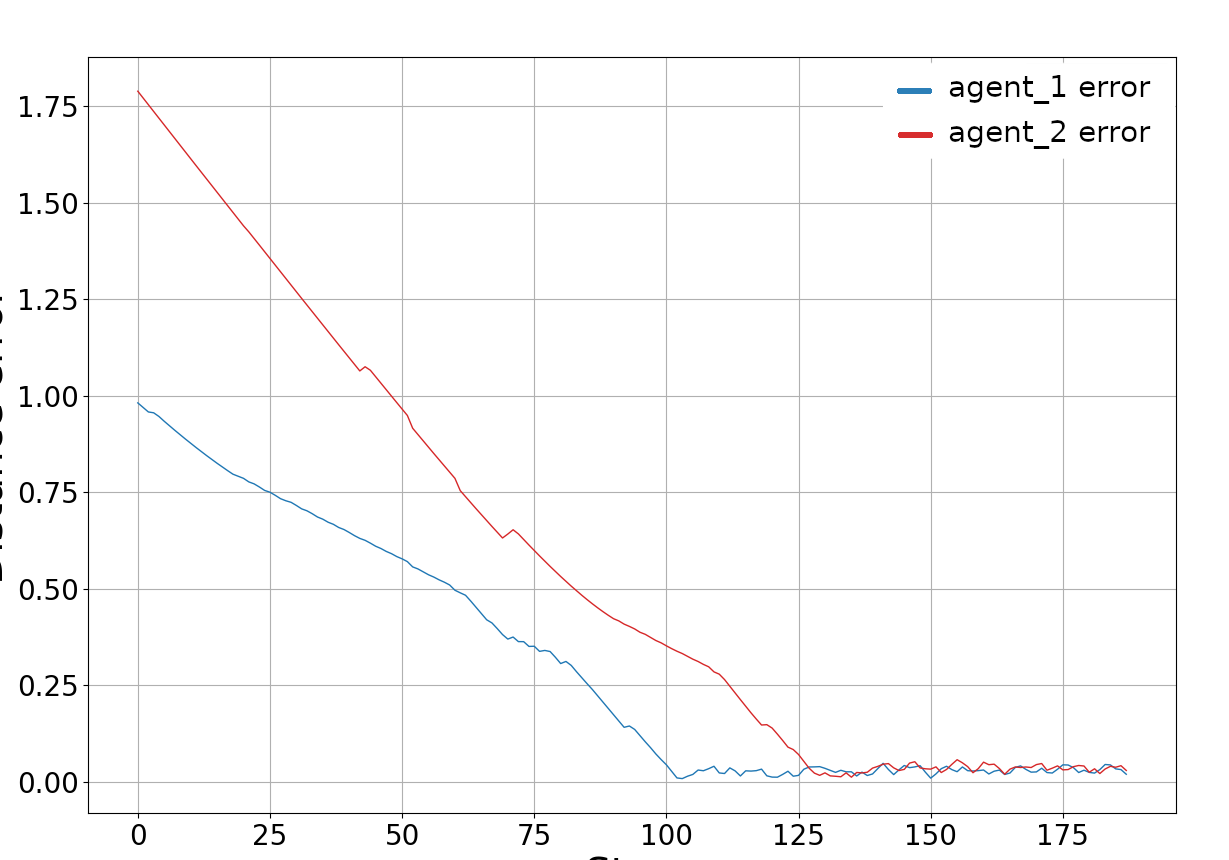}
    \caption{Distance error calculated in test shown on Fig.~\ref{fig:simulation-is4}}
    \label{fig:disterr}
\end{figure}

In order to test the behavior of the obstacle avoidance algorithm, a setup consisting of three agents was employed. The leader agent was maintaining its position, while the other two agents were required to move towards a predetermined triangular position. Through the use of four distinct setups, as shown in Fig.~\ref{fig:simulation-resultsReaching}, it was observed that the agents were successful in avoiding potential obstacles within their flock, ultimately resulting in a secure approach to the target position. It was observed that the agents displayed higher confidence levels when situated at a greater distance from the target, as they tended to approach it in a gradual manner as they moved closer. This may be attributed to the fact that the model was trained using a scenario in which the leader agent was in random motion, and thus, the agents were trained to maintain a safe distance from the leader while slowly approaching the target. To address this issue, it may be possible to incorporate constant negative rewards at each step, though this could potentially lead to an increase in collision rates. Alternatively, incorporating the velocity of the leader agent into the model's state may result in a more optimal approach stage while still maintaining low collision rates.
In Fig.~\ref{fig:disterr}, the distance error of two agents is shown, indicating that both agents initially moved towards the target, reducing the distance error. As the target was approached, the agents were able to maintain a distance error close to 0.

\begin{figure*}[t]
    \centering
    \begin{subfigure}[t]{0.3\textwidth}
        \centering
        \includegraphics[width=\textwidth]{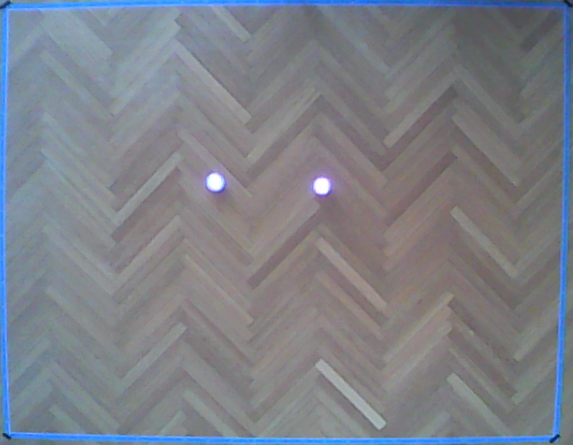}
        \caption{Camera view of two Sphero robots in the experimental environment}
        \label{fig:undetected}
    \end{subfigure}
    \hfill
    \begin{subfigure}[t]{0.3\textwidth}
        \centering
        \includegraphics[width=\textwidth]{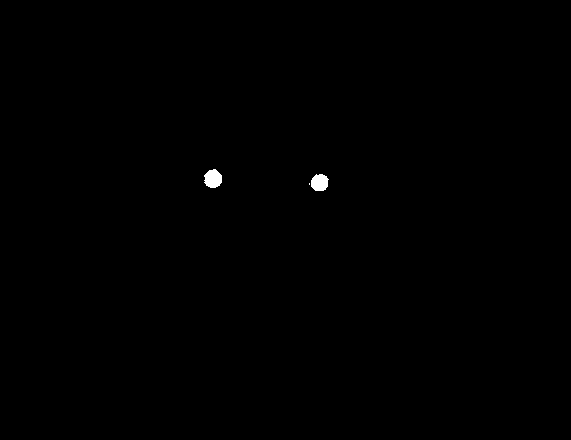}
        \caption{Mask used for object detection}
        \label{fig:mask}
    \end{subfigure}
    \hfill
    \begin{subfigure}[t]{0.3\textwidth}
        \centering
        \includegraphics[width=\textwidth]{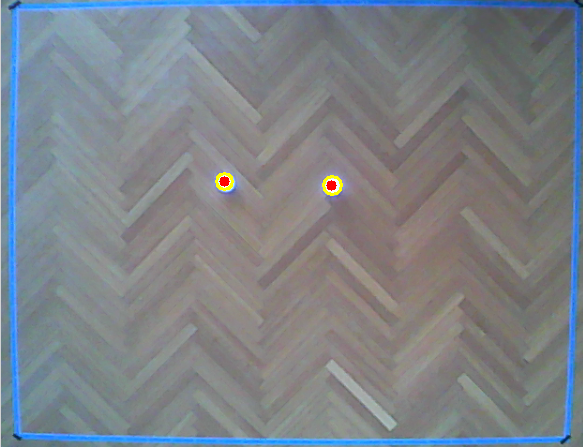}
        \caption{Two detected Sphero robots in the environment}
        \label{fig:detected}
    \end{subfigure}
    \caption{Detection pipeline}
    \label{fig:localization} 
\end{figure*}

\subsection{Real-world experiments}
The experimental setup consists of an HD web camera for global localization, Sphero robots, and a 2.5 m x 2 m environment. The view from the web camera can be seen in  Fig.~\ref{fig:undetected}.
Spheros' white LEDs allow us to detect them in the environment easily using a simple brightness threshold mask (Fig.~\ref{fig:mask}). Erosion and dilation are applied to the mask to remove any small blobs caused by image artifacts. Finally, the biggest contours are selected using the OpenCV library~\cite{opencv_library}.
Fig.~\ref{fig:detected} displays two detected Sphero robots in the environment. The robot's center is shown as a red dot and the robot's edge as a yellow circle. Robot detection proved to work perfectly, and robots were detected easily in every frame. 
Once the detections are available, robots are tracked with a visual multiple object tracking algorithm SORT~\cite{Bewley2016_sort}. Each robot is now placed inside the bounding box and has an assigned index. Finally, the Kalman filter is used to further enhance the robustness of localization.

\begin{figure}[t]
    \centering
    \begin{subfigure}[t]{0.48\columnwidth}
        \centering
        \includegraphics[width=\textwidth]{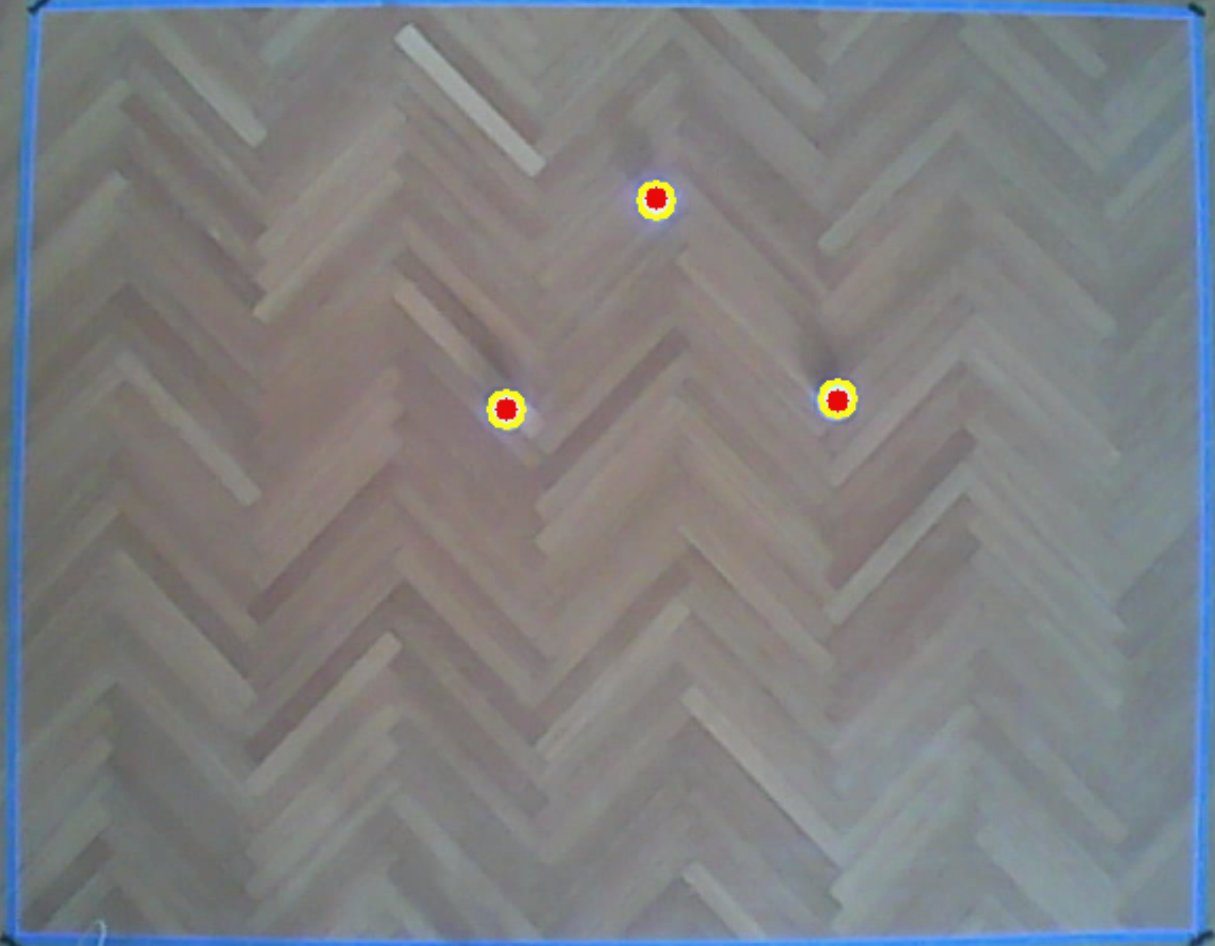}
        \caption{Initial position}
        \label{fig:sub-first4}
    \end{subfigure}
    \hfill
    \begin{subfigure}[t]{0.48\columnwidth}
        \centering
        \includegraphics[width=\textwidth]{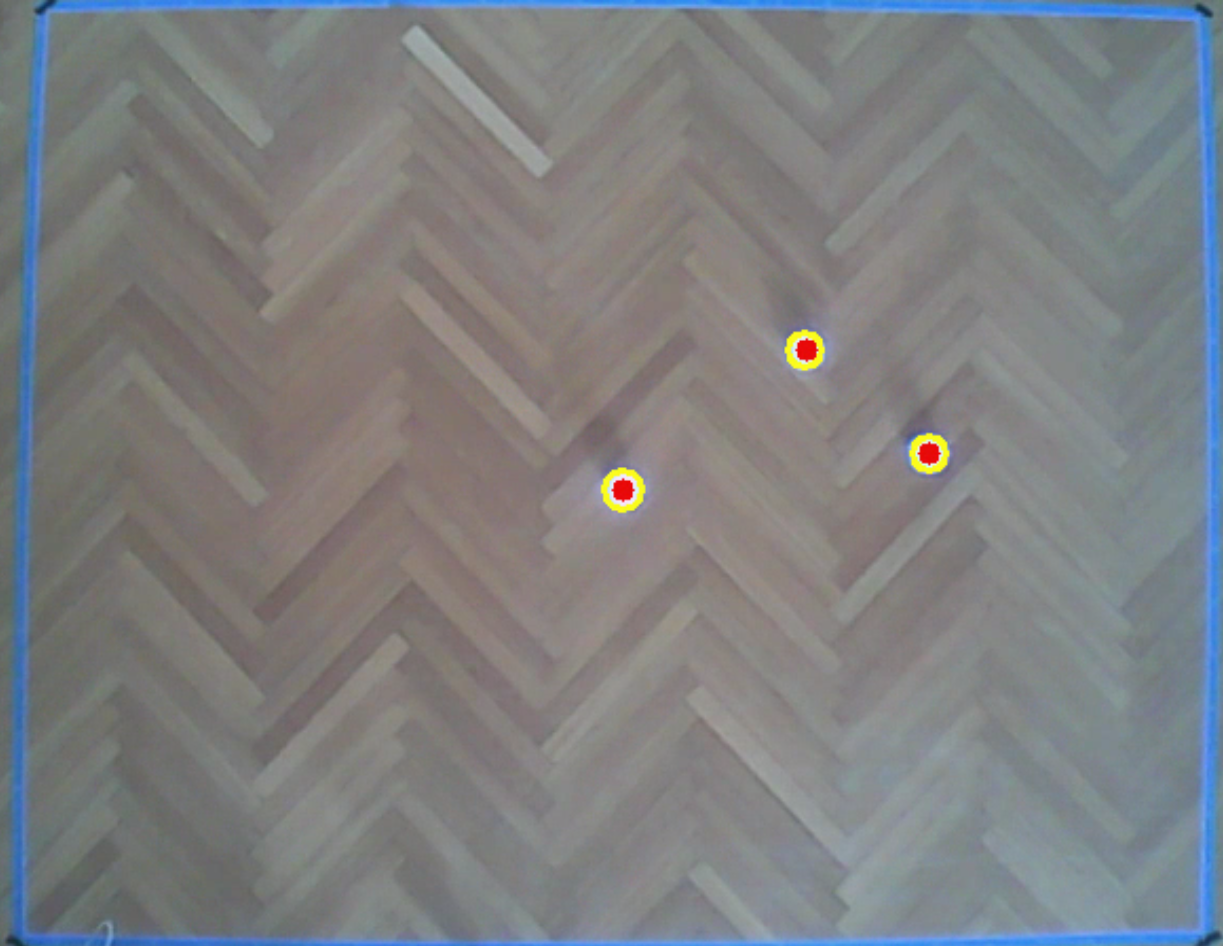}
        \caption{Formation starts to move}
        \label{fig:sub-second4}
    \end{subfigure}
    \\
    \begin{subfigure}[t]{0.48\columnwidth}
        \centering
        \includegraphics[width=\textwidth]{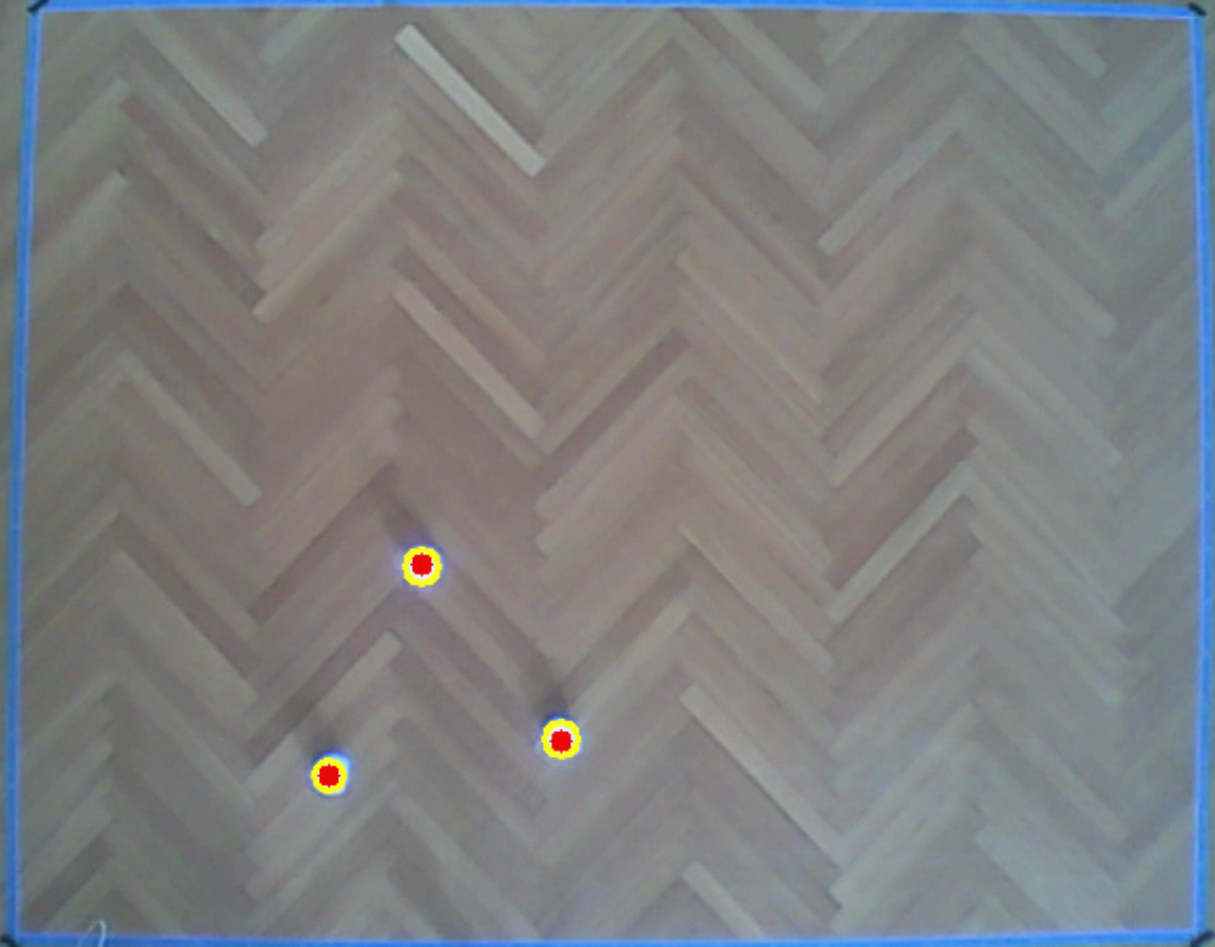}
        \caption{Formation moving in a circular motion}
        \label{fig:sub-third4}
    \end{subfigure}
    \hfill
    \begin{subfigure}[t]{0.48\columnwidth}
        \centering
        \includegraphics[width=\textwidth]{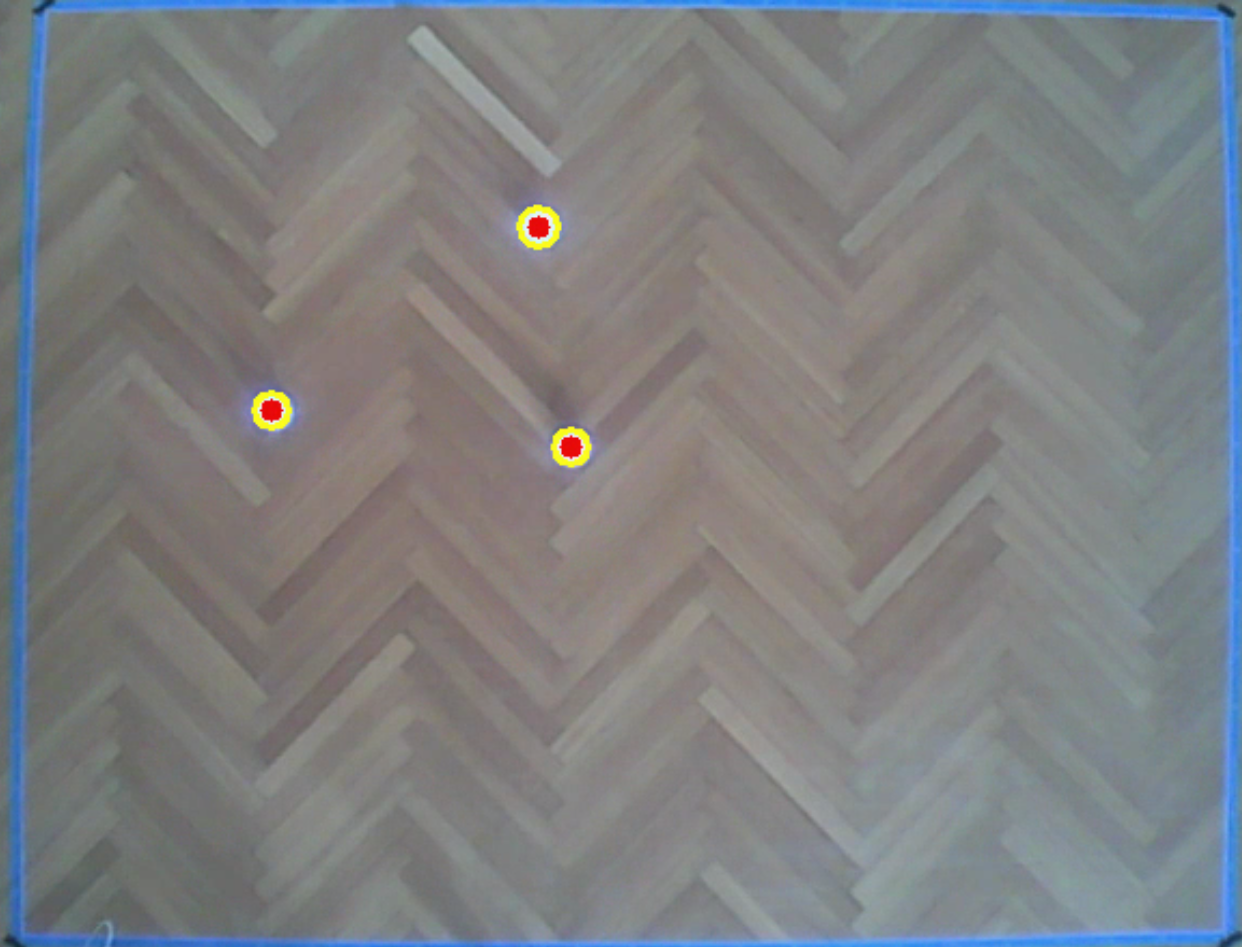}
        \caption{Formation almost finished one circle}
        \label{fig:sub-fourth4}
    \end{subfigure}
    \caption{Formation keeping experiment}
    \label{fig:formation-keeping} 
\end{figure}

\begin{figure}[hb]
    \centering
    \includegraphics[width=0.95\columnwidth]{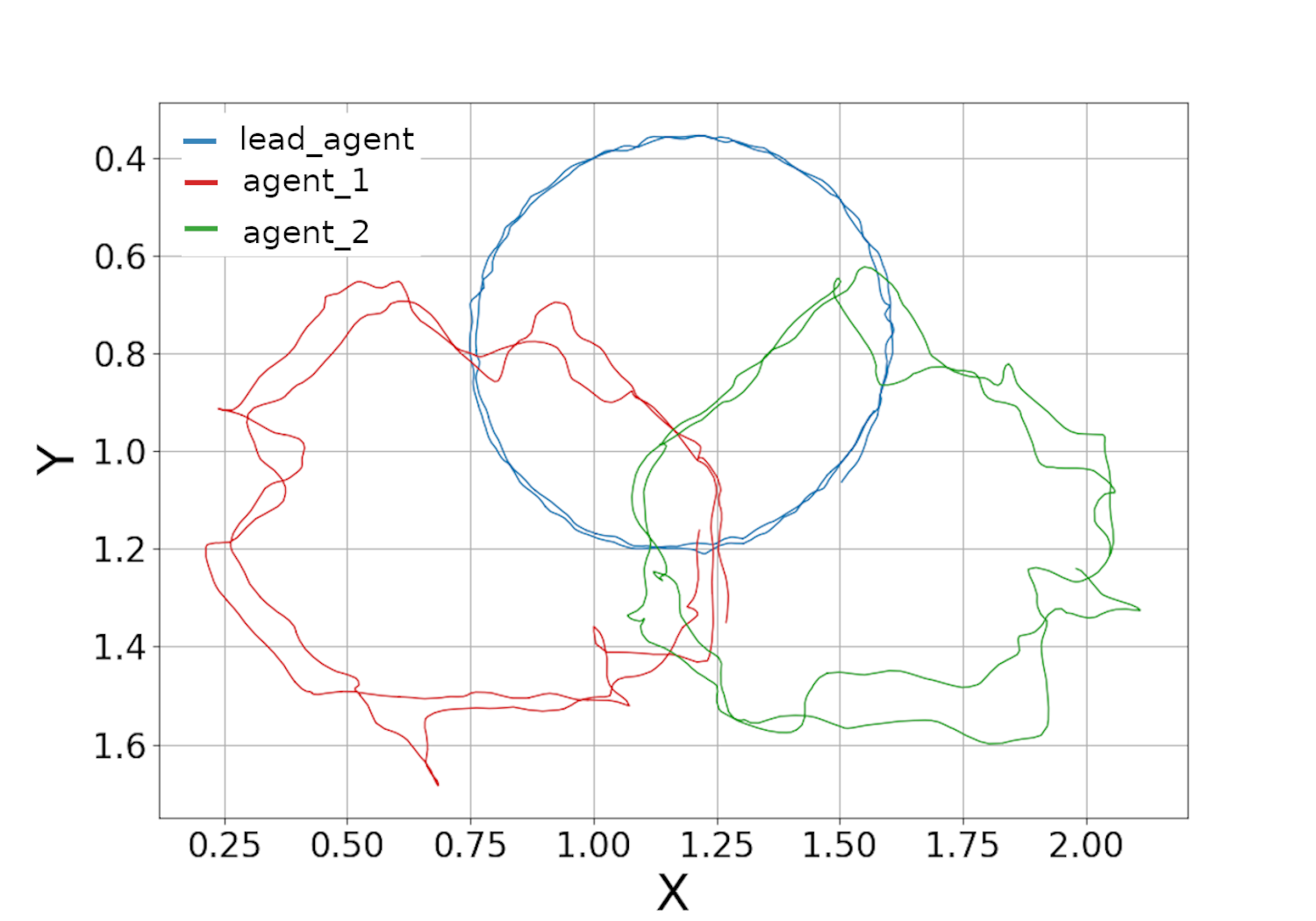}
    \caption{Robot paths in the real experiment}
    \label{fig:pathexp}
\end{figure}

Fig.~\ref{fig:formation-keeping} displays an experiment where the lead agent is moving in a circular motion and two other agents have to keep their position. Paths of the three agents in the experiment are shown in Fig.~\ref{fig:pathexp}.
Compared with the simulation, it is obvious that experimental results are much more unstable. This can be attributed to different environmental factors, such as robot inertia, which was not considered in the simulation, sensor errors, and delay in communication. However, both agents successfully follow the leader without collisions. 

\section{Conclusion}
\label{sec:conclusion}
In this paper we aimed to develop a decentralized algorithm for leader-follower formation control in a multi-robot system, utilizing the DDQN algorithm.  The results showed that by separating the problem into two simpler tasks we were able to accomplish safe navigation of the robot in the approach phase and precise and consistent position-keeping using a simple learning-based method. Algorithm was tested in the simulation phase as well in the real-world experiment setup. To improve algorithm stability in the experimental phase, future work will focus on incorporating the velocity of other agents into the model's state. Furthermore, the algorithm will also be implemented using different reinforcement learning algorithms to compare the performance with the currently used DDQN algorithm.

\addtolength{\textheight}{-6cm}   
\bibliographystyle{IEEEtran}
\bibliography{references_new.bib}  

\end{document}